\documentclass[NETN,manuscript]{stjour-new}

\newcommand{\ie}{\textit{i.e., }}
\newcommand{\eg}{\textit{e.g., }}

\articletype{Research}

\graphicspath{{figs/}}
\usepackage{array}
\usepackage{diagbox}
\usepackage{float}
\usepackage{bbm}
\usepackage{graphicx}
\usepackage{amsmath,amssymb}
\usepackage{adjustbox}
\usepackage{caption}
\usepackage{subcaption}
\usepackage{multirow}
\usepackage{url}
\usepackage{float}
\usepackage{cite}
\usepackage{subfloat}
\usepackage{pifont}
\usepackage{color}
\usepackage{xcolor}
\usepackage{soul}
\usepackage{booktabs}

\usepackage{makecell} 
\usepackage[margin=1in]{geometry}
\usepackage{graphicx}
\usepackage{booktabs}
\usepackage{caption}
\usepackage{adjustbox}

\newcolumntype{C}[1]{>{\centering\arraybackslash}m{#1}}
\usepackage{longtable}

\begin{document}
\title{Predicting Cognition from fMRI: \\
A Comparative Study of Graph, Transformer, and Kernel Models \\
Across Task and Rest Conditions} 

\author[Author Names]
{Jagruti Patel\affil{1}, Mikkel Schöttner\affil{1}, Thomas A. W. Bolton\affil{1}, Patric Hagmann\affil{1}}

\affiliation{1}{Department of Radiology, Lausanne University Hospital and University of Lausanne (CHUV-UNIL), Lausanne, Switzerland}

\correspondingauthor{Jagruti Patel}{jagruti90@gmail.com}

\vspace{4em}

\begin{abstract}

Predicting cognition from neuroimaging data in healthy individuals offers insights into the neural mechanisms underlying cognitive abilities, with potential applications in precision medicine and early detection of neurological and psychiatric conditions. This study systematically benchmarked classical machine learning (Kernel Ridge Regression) and advanced deep learning models (Graph Neural Networks and Transformer-GNNs) for cognitive prediction using Resting-state, Working Memory, and Language task fMRI data from the Human Connectome Project Young Adult (HCP-YA) dataset.

Our results, based on $R^2$ scores, Pearson correlation coefficient, and mean absolute error, revealed that task-based fMRI, eliciting neural responses directly tied to cognition, outperformed Resting-state fMRI in predicting cognitive behavior. Among the methods compared, a GNN combining structural and functional connectivity consistently achieved the highest performance across all fMRI modalities; however, its advantage over Kernel Ridge Regression using functional connectivity alone was not statistically significant. The Transformer-GNN, designed to model temporal dynamics with structural connectivity as a prior, performed competitively with FC-based approaches for task-fMRI but struggled with Resting-state data, where its performance aligned with the lower-performing GNN that directly used fMRI time-series data as node features. These findings emphasize the importance of selecting appropriate model architectures and feature representations to fully leverage the spatial and temporal richness of neuroimaging data.

This study highlights the potential of multimodal graph-aware deep learning models to combine structural and functional connectivity for cognitive prediction, as well as the promise of Transformer-based approaches for capturing temporal dynamics. By providing a comprehensive comparison of models, this work serves as a guide for advancing brain-behavior modeling using fMRI, structural connectivity and deep learning.  

\end{abstract}


\newpage

\newpage

\section{Introduction}

Understanding and predicting behavior from neuroimaging data in healthy individuals is crucial for advancing our knowledge of the brain's functional architecture and its relationship to behavior. While significant efforts have focused on patients with neurological or psychiatric disorders \citep{sabuncu2015clinical, arbabshirani2017single}, the study of healthy participants remains underexplored. Analyzing brain connectivity in healthy individuals can provide valuable insights into the baseline neural mechanisms underlying behavior, offering a foundation for early prognosis of potential neuro or psychiatric conditions \citep{zhou2012predicting, fornito2015connectomics, bassett2017network, lui2016psychoradiology}. By examining the intricate patterns of functional and structural connectivity, we can identify biomarkers indicative of brain health, which can serve as early indicators of disease susceptibility \citep{kaiser2013potential, li2023individualized}. This approach not only enhances our understanding of normal brain function but also aids in developing targeted interventions and precision medicine strategies, ultimately improving treatment outcomes and quality of life \citep{bzdok2018machine, dubois2016building, finn2016individual}.

To address these needs, machine learning (ML) has been pivotal in predicting behavior from neuroimaging data, offering insights into the complex relationships between brain structure, function, and behavior. Functional connectivity (FC) from resting-state or task-based fMRI is particularly valued for capturing neural synchrony across brain regions, outperforming other modalities like anatomical or diffusion MRI or structural connectivity (SC), especially for cognitive predictions \citep{ooi2022comparison, dhamala2021distinct, tian2021high}. Classical ML techniques, such as kernel ridge regression, linear ridge regression, and elastic net, have shown effectiveness in handling high-dimensional data, providing robust performance even with smaller sample sizes \citep{ooi2022comparison, he2020deep}. However, the advent of deep learning (DL) models, including deep neural networks (DNNs), convolutional neural networks (CNNs), and graph neural networks (GNNs), offers significant potential due to their ability to capture complex, non-linear patterns and extract hierarchical features from large datasets \citep{thapaliya2025brain,xia2025interpretable,wen2024multi,huang2022spatio,hanik2022predicting,jegham2022meta,he2020deep}). Despite the challenges posed by neuroimaging data, such as noise, inter-subject variability, and the intricate relationship between brain structure and behavior, DL methods offer a promising path to advance predictive accuracy and deepen our understanding of neural underpinnings. Spatio-temporal models, for instance, integrate dynamic temporal and spatial information, further enhancing DL's capabilities \citep{huang2022spatio}. With the increasing availability of large-scale datasets, it is crucial to thoroughly investigate and explore DL methods to fully exploit their potential in revealing complex brain-behavior relationships \citep{vieira2024beyond, vieira2017using}.

FC derived from resting-state fMRI has become a cornerstone in behavior modeling due to its ability to capture synchronous neural activity across brain regions, offering valuable insights into the brain's functional organization \citep{biswal1995functional,fox2007spontaneous,ooi2022comparison}. It has been instrumental in predicting behavioral traits by analyzing inter-regional interactions \citep{finn2015functional,smith2015positive,he2020deep}. While resting-state FC offers a robust baseline for predicting behavioral dimensions like cognition, emerging evidence suggests that FC derived from task-fMRI, particularly when the tasks are designed to engage relevant cognitive processes, can yield superior predictive power \citep{zhao2023task, greene2018task, sripada2020toward, kraljevic2024network}. Conversely, task-based fMRI acquired during tasks unrelated to the cognitive domain being modeled has been shown to underperform compared to resting-state FC \citep{ooi2022comparison}. This highlights the inherent utility of resting-state FC as a general predictor alongside the potential of carefully designed tasks to evoke more targeted neural activity patterns relevant to specific behaviors.
While FC, both at rest and during tasks, has shown promise, recent advancements in DL have introduced sophisticated architectures like Graph-Transformers for modeling the intricate relationships within FC data and predict cognitive abilities \citep{qu2024exploring, qu2023interpretable, hussain2024rct}.  

To move towards a more comprehensive understanding of the neural substrates of behavior, multimodal approaches leveraging complementary information from different data modalities are gaining prominence. These integrated strategies, have been shown to improve the prediction of behavior \citep{rasero2021integrating, kramer2024prediction, xia2025interpretable, qu2025integrated}. Specifically, incorporating both SC and FC knowledge boosts the predictive power of behavior by capturing compensatory mechanisms and inter-regional interactions \citep{kramer2024prediction, xia2025interpretable, dhamala2020integrating, dsouza2021m}. This enhanced understanding of brain network architecture through the combined lens of different modalities leads to more accurate and robust modeling of behavioral predictions, which is crucial for advancing both basic neuroscience and clinical applications.

Building on the need for comprehensive modeling, working directly with fMRI time-series data, rather than relying solely on derived FC matrices, offers the potential to capture richer information about dynamic interactions and temporal dependencies in neural activity \citep{vieira2024beyond}. Time-series data preserves the temporal order and variability of neural signals, crucial for understanding complex, time-varying patterns associated with cognitive traits like \citep{fan2020deep}. Recent studies have explored this approach using models like recurrent neural networks (RNNs), LSTMs or Transformers to predict behavior directly from regional time-series \citep{hebling2021deep, li2023accounting, kim2023swift}. These methods aim to leverage temporal dynamics that might be lost in static FC representations, allowing for a more nuanced characterization of brain function. However, many existing time-series methods struggle to effectively integrate spatial and temporal information, which is essential for capturing the full scope of brain-behavior relationships \citep{fan2020deep}. Spatio-temporal modeling addresses this limitation by simultaneously capturing the temporal dynamics and spatial structure of brain networks, offering a more comprehensive framework for behavior modeling \citep{huang2022spatio, yan2022modeling}. This approach can enhance the predictive power of neuroimaging analyses by fully exploiting the rich data available from fMRI studies.

In this context, our paper aims to benchmark a range of methods, from classical ML to advanced DL techniques, for predicting cognitive behavior using three types of fMRI time-series data: resting-state, working memory task, and language task.
Specifically, we evaluate four categories of methods:
(1) \textbf{Kernel Ridge Regression} (KRR) for its demonstrated effectiveness with FC \citep{he2020deep}, providing a robust baseline due to its ability to handle high-dimensional data efficiently;
(2) \textbf{GNNs integrating FC and SC} to leverage the complementary strengths of functional and structural data, capturing both synchronous neural activity and the physical pathways that underpin these interactions \citep{xia2025interpretable};
(3) \textbf{GNNs incorporating SC and fMRI time-series data} to enhance model representation, allowing for a more nuanced understanding of temporal patterns alongside structural connectivity \citep{fan2020deep, hebling2021deep}; and
(4) a \textbf{Spatio-Temporal Transformer-GNN model} that uses a Transformer to encode temporal dynamics and a GNN to capture spatial relationships, offering a comprehensive framework that fully exploits both temporal and spatial information.
Benchmarking these methods is crucial to understand their potential for advancing behavior prediction. By evaluating models of increasing complexity and information richness, we aim to identify those offering the most significant improvements in predictive power. This study uses the Human Connectome Project Young Adult (HCP-YA) \citep{van2012human, van2013wu, glasser2013minimal, sotiropoulos2013advances} dataset to regress cognitive behavior from the three types of fMRI time-series data, as cognition prediction has been empirically shown to be more predictive than other behavioral prediction tasks, such as dissatisfaction and emotion \citep{ooi2022comparison}. Through this comprehensive comparison, we seek to guide future research in developing more effective models for behavioral and, specifically, cognitive prediction.
 
\section{Dataset and Preprocessing}

Minimally preprocessed T1-weighted, diffusion-weighted, and both resting-state and task-based fMRI data was obtained from HCP-YA dataset. Cognitive behavioral scores were used from \citep{ooi2022comparison}. Only subjects with complete structural and functional imaging data, as well as behavioral assessments, were retained—resulting in a final sample of 690 individuals (ages 22–37, 367 females and 323 males).

SC matrices were computed using Connectome Mapper 3 (CMP3) version 3.0.0-rc4 \citep{tourbier2022connectome, tourbier2020connectomicslab} from T1-weighted and diffusion-weighted images. For computational efficiency, T1-weighted images were downsampled from 0.7 mm to 1.25 mm isotropic resolution. Single-shell diffusion-weighted images with a b-value of 3000 s/mm\textsuperscript{2} and 1.25 mm isotropic resolution were used for tractography.
T1-weighted images were parcellated using FreeSurfer’s \texttt{recon-all} pipeline (version 6.0.1) with the Lausanne Scale 3 atlas \citep{cammoun2012mapping, iglesias2015computational, iglesias2015bayesian, najdenovska2018vivo, desikan2006automated}, resulting in 274 cortical and subcortical regions. Tractograms were generated from diffusion weighted images using MRtrix3 \citep{tournier2012mrtrix} with constrained spherical deconvolution of order 8 \citep{tournier2007robust}, white matter seeding, and deterministic tracking to generate 10 million streamlines. These tractograms were then combined with the parcellated T1-weighted images to produce strutural connectivity matrices, where edge weights represent the number of fibers connecting each pair of brain regions (with self-connections set to zero).

fMRI data had already undergone HCP’s minimal preprocessing pipeline, including distortion correction, realignment, co-registration to structural images, bias field correction, normalization, and brain masking \citep{glasser2013minimal}. Additional preprocessing steps included discarding the first six volumes of the fMRI time-series data to account for scanner drift, followed by motion and nuisance regression using six head motion parameters and their first-order derivatives, along with average white matter and cerebrospinal fluid signals. Voxel-wise detrending and high-pass filtering (0.01 Hz cutoff) were also applied. All these steps were performed using Nilearn’s \texttt{clean\_img} function.
To extract region-wise fMRI time-series, all data were aligned to a common space. For each subject, the Lausanne Scale 3 parcellation (originally in native space) was warped to Montreal Neurological Institute  space using FSL’s \texttt{applywarp} function with nearest-neighbor interpolation. Functional time-series were then extracted by averaging voxel time courses within each parcel using Nilearn’s \texttt{NiftiLabelsMasker.fit\_transform()}. These time-series were z-score normalized across time, and FC matrices were computed as the Pearson correlation coefficient between every pair of parcellated regional time courses.

To derive behavioral scores, \citep{ooi2022comparison} applied Principal Component Analysis to 58 behavioral measures from a separate cohort, independent of the main study subjects. The top three components explaining the largest proportion of variance were retained and subsequently subjected to Varimax rotation \citep{kaiser1958varimax} to improve interpretability. Based on the resulting factor loadings, one of the components was identified as reflecting cognitive function. Using this factor structure, cognition-related factor scores were then estimated for the subjects included in the main study.

The dataset was split using 10-fold cross-validation, carefully controlling for family structure to prevent data leakage—ensuring that all family members were assigned to the same fold \citep{ooi2022comparison, vieira2024beyond}. Additionally, the folds were stratified to maintain a balanced distribution of the target behavioral variable (i.e., cognition scores) across folds. It ensured that the training set represented the overall distribution, and the validation set is reflected the test set distribution.
Each fold served once as the validation set and once as the test set over the course of cross-validation. Prior to model training, cognition scores were adjusted for age and gender via linear regression fitted on the training data; the resulting residualization model was then applied to both the validation and test sets. For DL models, cognition scores were further standardized to have zero mean and unit variance using the training set statistics. During inference, predicted scores were inverse-transformed to recover values on the original scale.


\section{Methods}

\subsection{Notations}
In this study, let the dataset $\mathcal{D}$ consists of information about $M$ subjects. 
For a subject with $N$ regions of interest (RoIs), the structural connectivity (SC) is denoted as $\mathbf{S} \in \mathbb{R}^{N \times N}$.
Let the fMRI time-series data for the subject is given as $\mathbf{H} \in \mathbb{R}^{N \times d}$ matrix, where $d$ is the number of measured time points.
Thereby, the FC is represented as $\mathbf{F} \in \mathbb{R}^{N \times N}$ matrix, which is computed as the pairwise Pearson correlation coefficients among the RoIs from  $\mathbf{H}$. Let $y$ denote the cognitive behavioral score, 
for the subject. 
Thus, $\mathcal{D}$ is defined as $\{\mathcal{S}, \mathcal{F}, \mathcal{H}, \mathcal{Y}\}$, where $\mathcal{S}: \{\mathbf{S}_i\}_{i=1:M}$, $\mathcal{F}: \{\mathbf{F}_i\}_{i=1:M}$, $\mathcal{H}: \{\mathbf{H}_i\}_{i=1:M}$, and $\mathcal{Y}: \{y_i\}_{i=1:M}$.

\subsection{Kernel Ridge Regression}
Kernel Ridge Regression (KRR) is used to predict the behavioral score $y_i$ for a subject $i$ using the subject's brain features from $\mathbf{F}_i$ (\eg, lower triangle of the FC matrix). For the training set with $M$ subjects, the KRR model expresses $y_i$ as:
\[
y_i = \beta_0 + \sum_{j=1}^M \alpha_j K(\mathbf{F}_j, \mathbf{F}_i)
\]
where $\beta_0$ is the bias term, $\alpha_j$ are model weights, and $K(\mathbf{F}_j, \mathbf{F}_i)$ represents the similarity between the brain features $\mathbf{F}_j$ and $\mathbf{F}_i$ of subjects $j$ and $i$, respectively. In this study, $K(\mathbf{F}_j, \mathbf{F}_i)$ is calculated as the dot product
between the vectorized forms of $\mathbf{F}_j$ and $\mathbf{F}_i$.

To estimate the parameters $\boldsymbol{\alpha} = [\alpha_1, \alpha_2, \dots, \alpha_M]^\top$ and $\beta_0$, we define $\mathbf{y} = [y_1, y_2, \dots, y_M]^\top$ and $\kappa$ as an $M \times M$ similarity matrix where the $(j, i)$-th element is $K(\mathbf{F}_j, \mathbf{F}_i)$. These parameters are optimized by minimizing the $L_2$-regularized cost function, given as,
\[
(\boldsymbol{\alpha}, \beta_0) = \arg\min_{\boldsymbol{\alpha}, \beta_0} \frac{1}{2} \|\mathbf{y} - \kappa \boldsymbol{\alpha} - \beta_0\|^2 + \frac{\lambda}{2} \boldsymbol{\alpha}^\top \kappa \boldsymbol{\alpha}
\]
where $\lambda$ is the regularization parameter that controls overfitting and is determined via the inner-loop of nested-cross-validation procedure using the training data.
For a new subject $t$, the predicted behavioral score $\hat{y}_t$ is computed as:
\[
\hat{y}_t = \beta_0 + \sum_{i=1}^M \alpha_i K(\mathbf{F}_i, \mathbf{F}_t),
\]
where $\mathbf{F}_t$ is the FC matrix of the test subject, and $K(\mathbf{F}_i, \mathbf{F}_t)$ measures its similarity to the training subjects' FC matrices.

\subsection{Preliminaries of Deep Learning Methods}
This section introduces the foundational concepts—GNNs and Transformers—crucial for understanding the deep learning-based methods discussed in Section~\nameref{sec:dl_modeling}.

\textbf{\textit{Graph Neural Networks (GNNs)}:}
GNN is used to predict the behavioral score $y_i$ for a subject $i$ using the corresponding SC matrix $\mathbf{S}_i$, $\mathbf{F}_i$, and $\mathbf{H}_i$. GNNs are employed to learn a richer representation of the underlying patterns, incorporating the spatial and temporal relations among the RoIs, that drive behavior dynamics. Specifically in this study, the GNNs are based on Chebyshev spectral graph convolutions as they provide an efficient approach for processing graph-structured data by leveraging local spectral information. 

Let $\mathbf{G}_i: \{\mathbf{V}_i, \mathbf{A}_i\}$ denotes the graph representation for subject $i$, where $\mathbf{V}_i \in \mathbb{R}^{N \times d}$ and $\mathbf{A}_i \in \mathbb{R}^{N \times N}$ denote the node feature matrix and the symmetric weighted adjacency matrix, respectively. $N$ denotes the number of nodes or the RoIs. For the ease of readability, we are dropping $i$ in the following equations. The GNN uses a Chebyshev graph convolution (ChebConv) layer with polynomial order $K$, which allows each node to aggregate information from its $K$-hop neighborhood. This enables the model to capture localized patterns up to $K$ hops away in the graph structure.


A ChebConv layer begins with computing the normalized graph Laplacian $\mathbf{L}$ as follows where, $\mathbf{I}_N$ is a $N \times N$ identity matrix and $\mathbf{D}$ is the diagonal degree matrix with elements $\mathbf{D}_{uu} = \sum_{v=1}^N \mathbf{A}_{uv}$. 
\[
\mathbf{L} = \mathbf{I}_N - \mathbf{D}^{-\frac{1}{2}} \mathbf{A} \mathbf{D}^{-\frac{1}{2}}
\]
The Laplacian $\mathbf{L}$ can be decomposed as $\mathbf{L} = \mathbf{U} \mathbf{\Lambda} \mathbf{U}^\top$, where $\mathbf{U} \in \mathbb{R}^{N \times N}$ is the matrix of eigenvectors and $\mathbf{\Lambda} = \text{diag}([\lambda_1, \lambda_2, \dots, \lambda_N])$ is a diagonal matrix containing the corresponding eigenvalues. 

A signal $\mathbf{x} \in \mathbb{R}^N$ defined on the graph's vertices can be transformed into the spectral domain as $\hat{\mathbf{x}} = \mathbf{U}^\top \mathbf{x}$. A convolution operation on $\mathbf{x}$ with a graph filter $g$ in the vertex domain is equivalent to multiplication in the spectral domain as, 
\[
\mathbf{x} * g = \mathbf{U} \hat{g}(\mathbf{\Lambda}) \mathbf{U}^\top \mathbf{x} = \hat{g}(\mathbf{L}) \mathbf{x}
\]
Here, $\hat{g}(\mathbf{\Lambda})$ denotes the filter function applied to the eigenvalues of $\mathbf{L}$, and $\hat{g}(\mathbf{L})$ is its matrix polynomial form. To ensure that $\hat{g}(\mathbf{L})$ is $K$-localized (\ie, dependent only on nodes up to $K$ hops away) and to avoid the computational cost of eigen-decomposition, it is approximated using Chebyshev polynomials of order $K$, given as,
\[
g_\theta(\mathbf{L}) = \sum_{k=0}^{K} \theta_k T_k(\tilde{\mathbf{L}})
\]
where, 
$T_k(x)$ are Chebyshev polynomials of degree $k$,
$\tilde{\mathbf{L}} = \frac{2}{\lambda_{\max}} \mathbf{L} - \mathbf{I}_N$ is the rescaled graph Laplacian,
$\lambda_{\max}$ is the largest eigenvalue of $\mathbf{L}$, and
$\theta_k$ are the trainable parameters of the GNN.
The Chebyshev polynomials \( T_k(x) \) are defined recursively as:
\[
T_0(x) = 1, \quad T_1(x) = x, \quad T_k(x) = 2x T_{k-1}(x) - T_{k-2}(x), \quad k \geq 2
\]
Using the above formulation and approximation, a ChebConv layer updates the node features $\mathbf{X}^{(k)} \in \mathbb{R}^{N \times d^{(k)}}$ from layer $k$ to layer $k+1$ as, 
\[
\mathbf{H}^{(k+1)} = \sigma \left( \text{Norm} \left( \sum_{k=0}^{K} \theta_k T_k(\tilde{\mathbf{L}}) \mathbf{H}^{(k)} \right) \right)
\]
where, $d^{(k)}$ is the feature dimension at layer $k$,
$\sigma$ is a non-linear activation such as ReLU, and
\text{Norm} denotes a normalization operation (\eg, batch normalization). 

Following $K_{N}$ GNN layers, 
an order invariant pooling operation, \eg sum, mean, is employed to aggregate the node features for defining a graph-level representation. Subsequently, an MLP maps the graph-level representation to the output behavior score $y$.

\textbf{\textit{Transformers}:}
Transformer is a versatile architecture designed to model global dependencies in data using self-attention mechanisms. In this study, we leverage a Transformer to learn temporal representations from fMRI time-series data by treating each brain region $n \in N$ as a token and its associated time-series as input features. Specifically, for each subject $i \in M$, the time-series matrix $\mathbf{H} \in \mathbb{R}^{N \times d}$—with $N$ brain regions and $d$ time points—is projected to $\mathbf{H}' \in \mathbb{R}^{N \times d'}$, where $d'<d$. As defining temporal relations among the measurements is infeasible, we aim to learn them in a data-driven manner via a Transformer. A generic formulation of Transformers is presented below.


Given an input $\mathbf{X} \in \mathbb{R}^{T \times q}$ with $T$ tokens and token dimension $q$, the model begins with prepending a special learnable \texttt{[CLS]} token embedding $\mathbf{x}_{\texttt{[CLS]}} \in \mathbb{R}^q$ to $\mathbf{X}$. The \texttt{[CLS]} token supports global representation learning for downstream tasks, such as  behavior prediction. The updated input becomes, $\mathbf{X}' = [\mathbf{x}_{\texttt{[CLS]}}, \mathbf{X}]$, and $\mathbf{X}' \in \mathbb{R}^{(T+1) \times q}$. Further, $\mathbf{X}'$ is augmenting with positional encodings $\mathbf{P} \in \mathbb{R}^{(T+1) \times q}$. Considering the permutation-invariant nature of Transformers, $\mathbf{P}$ enables the model to encode relative and absolute positions of the input elements. Thus, the input to the Transformer becomes, $\tilde{\mathbf{X}} = \mathbf{X}' + \mathbf{P}$, where $\mathbf{P}$ is either learned or computed via fixed pre-defined functions. 

Using the augmented input $\tilde{\mathbf{X}}$, a Transformer layer computes queries ($\mathbf{Q}$), keys ($\mathbf{K}$), and values ($\mathbf{V}$) for subsequent attention computation, given as,
\[
\mathbf{Q} = \tilde{\mathbf{X}} \mathbf{W}_Q; \quad
\mathbf{K} = \tilde{\mathbf{X}} \mathbf{W}_K, \quad
\mathbf{V} = \tilde{\mathbf{X}} \mathbf{W}_V
\]
where $\mathbf{W}_Q, \mathbf{W}_K, \mathbf{W}_V \in \mathbb{R}^{q \times q_\text{head}}$ are learned projection matrices. $q_\text{head}$ is the dimensionality of each attention head $h$ in a Multi-Head Attention (MHA). The self-attention per head is computed as:
\[
\text{Attention}(\mathbf{Q}, \mathbf{K}, \mathbf{V}) = \text{softmax} \left( \frac{\mathbf{Q} \mathbf{K}^\top}{\sqrt{q_\text{head}}} \right) \mathbf{V},
\]
where $\sqrt{q_\text{head}}$ is a scaling factor for numerical stability. MHA concatenates and projects the outputs from multiple attention heads ensuring diverse interactions, given as,
\[
\text{MHA}(\mathbf{Q}, \mathbf{K}, \mathbf{V}) = \text{Concat} \left( \text{Attention}_1, \dots, \text{Attention}_h \right) \mathbf{W}_O
\]
where $\mathbf{W}_O \in \mathbb{R}^{(h \times q_\text{head}) \times q}$ is the learned output projection matrix. The output of MHA is processed by MLP, Layer normalization (LN), and residual connections, to complete one Transformer layer. Given $K$ Transformer layers and the input to the $k$-th layer as $\tilde{\mathbf{X}}^{(k)}$, a Transformer layer is summarized as,
\[
\tilde{\mathbf{X}}^{(k+1)} = \text{LN} \left( \tilde{\mathbf{X}}^{(k)} + \text{MHA}(\mathbf{Q}^{(k)}, \mathbf{K}^{(k)}, \mathbf{V}^{(k)} \right)
\]
\[
\tilde{\mathbf{X}}^{(k+1)} = \text{LN} \left( \tilde{\mathbf{X}}^{(k+1)} + \text{MLP}(\tilde{\mathbf{X}}^{(k+1)}) \right)
\]
After $K$ layers, the final output $\tilde{\mathbf{X}}^{(K)} \in \mathbb{R}^{(T+1) \times q}$ includes a summarized \texttt{[CLS]} token embedding $\mathbf{x}_{\texttt{[CLS]}}^{(K)}$ and $T$ contextualized token representations. The final \texttt{[CLS]} token can be used for downstream tasks.

\subsection{Deep Learning Methods}
\label{sec:dl_modeling}
In this subsection, we introduce four DL methods for predicting subject-specific behavioral scores. These methods primarily leverage GNNs, as GNNs offer a powerful framework for modeling brain connectivity, effectively capturing complex interactions among RoIs to enhance behavior prediction. 
For each subject $i$, a graph $\mathbf{G}_i$ is defined, where the adjacency matrix $\mathbf{A}_i$ is initialized using the SC matrix $\mathbf{S}_i$, which encodes anatomical pathways and establishes a biologically informed graph topology. This initialization incorporates fixed structural relationships, reflecting meaningful neural connections. While all methods share a consistent graph topology, they differ in defining the node feature matrices $\mathbf{V}_i$ for $\mathbf{G}_i$. Detailed descriptions of each method are provided in the following sections.


\textbf{\textit{GNNs with SC and FC Features}:}
\label{sec:gnn_fc}
In this method, the node feature matrix $\mathbf{V}_i$ is initialized using the FC matrix $\mathbf{F}_i$, where each row of $\mathbf{F}_i$ serves as a $1 \times N$ feature embedding for the corresponding node, capturing dynamic co-activation and inter-regional communication patterns derived from the Pearson correlation of fMRI time-series signals. By integrating the complementarities of SC (via $\mathbf{A}_i$) and FC (via $\mathbf{V}_i$), $\mathbf{G}_i$ enables the GNN to learn both structural and functional brain network properties effectively.  

The GNN employs $K_N$ ChebConv layers to iteratively propagate, aggregate, and update information across nodes, capturing higher-order dependencies and complex interactions shaped by the brain’s topology. Following the $K_N$ layers, updated embeddings are extracted for all $N$ nodes or RoIs.  
Subsequently, mean-pooling and max-pooling operations are performed on the node embeddings, with their outputs concatenated to generate graph-level representations $f_{\mathbf{G}_i}$, a global representation encapsulating the properties of the brain graph. To note, mean-pooling aggregates overall node activity and max-pooling highlights the most significant features across nodes.
Finally, $f_{\mathbf{G}_i}$ is passed through a multi-layer perceptron (MLP) regression head, mapping the aggregated information to the subject’s behavioral score $y_i$. By combining SC-driven adjacency topology, FC-driven node features, and the GNN’s dynamic integration capabilities, the approach efficiently captures the interplay between structural connections and functional co-activation patterns for robust behavior prediction.

\textbf{\textit{GNNs with SC and fMRI Time-series Features}:}
In this method, the node feature matrix $\mathbf{V}_i$ is initialized using the fMRI time-series feature matrix $\mathbf{H}_i$, defined for subject $i$. Using $\mathbf{H}_i$ as node features offers key advantages over the FC matrix $\mathbf{F}_i$.  
First, $\mathbf{H}_i$ preserves the temporal resolution and variability of brain activity, whereas $\mathbf{F}_i$ collapses time-series data into a static pairwise correlation matrix, losing critical temporal information. This preserved temporal resolution allows the model to capture finer-grained dynamics in brain activity that influence cognitive and behavioral outcomes.  
Second, with $\mathbf{H}_i$, the GNN can learn more complex, non-linear, and temporally sensitive patterns across nodes, enabling a richer and more flexible representation of functional dynamics.  

Following the graph initialization, the learning workflow mirrors that in Section~\nameref{sec:gnn_fc}. $K_N$ ChebConv layers dynamically update node embeddings based on SC topology, followed by pooling operations to derive a graph-level representation $f_{\mathbf{G}_i}$, which is passed to an MLP for regression. This method provides greater flexibility for the model to adaptively learn spatial dynamics via SC and temporal dynamics directly from fMRI data, eliminating reliance on pre-computed FC matrices.
It is important to note that while the greater flexibility allows for richer representations, it also increases susceptibility to noise in the temporal measurements, which may adversely impact the GNN’s learning performance.

\textbf{\textit{Transformer-GNN with SC and fMRI Time-series Features}:}
\label{sec:tgnn}
In this method, the node feature matrix $\mathbf{V}_i$ for each graph $\mathbf{G}_i$ is initialized using output embeddings from a Transformer. Nodes are treated as tokens with features $\mathbb{R}^{N \times K}$, where $N$ represents the number of RoIs and $K$ is the input feature dimension. These node features are passed through the Transformer, which models spatial and temporal relations among the nodes to generate rich, region-wise embeddings \textbf{TrfMRI} $\in \mathbb{R}^{N \times D}$, where $D$ is the output feature dimension.  
Subsequently, cosine similarity is applied to the Transformer’s output embeddings $\mathbb{R}^{N \times D}$ to compute a FC matrix $\mathbb{R}^{N \times N}$. Unlike conventional FC matrices, deterministically computed as pairwise Pearson correlations of direct time-series data, this Transformer-derived FC matrix is dynamic and spatio-temporally enriched, optimized to reflect patterns learned from the downstream prediction task.  

For behavior prediction, the workflow follows Section \textbf{GNNs with SC and FC Features}
The rows of the computed FC matrix define the node features $\mathbf{V}_i$, while the adjacency matrix $\mathbf{A}_i$ is defined using the SC matrix $\mathbf{S}_i$, completing the graph $\mathbf{G}_i$. This graph is processed by $K_N$ ChebConv layers, followed by pooling operations to compute graph-level representation $f_{\mathbf{G}_i}$, which is fed into an MLP regression head to predict behavioral scores $y_i$. By leveraging SC-based topology, fMRI-derived functional signals, and Transformer-learned embeddings, this approach effectively captures the interaction between structural organization and temporal dynamics for robust behavior prediction.  

A crucial component of this method is defining the node features $\mathbb{R}^{N \times K}$ to feed the Transformer. We design $K$ by concatenating three complementary components:
\begin{itemize}  
    \item \textbf{MLP-projected fMRI time-series features:} These features encode dynamic functional signals by projecting direct fMRI time-series data through an MLP, which extracts neural activation patterns critical for modeling temporal interactions.  
    

    \item \textbf{Structural eigenvector features from SC:} Derived from the spectral decomposition of the SC graph, these eigenvectors capture the global structural topology of the brain. Each node's (brain region's) representation in the first $k$ eigenvectors forms a spectral embedding that reflects its position in the graph’s spectral space. These embeddings provide insights into connectivity gradients and anatomical organization. The leading eigenvectors correspond to low-frequency modes, which emphasize smooth, long-range patterns and global connectivity structures across the brain.

    \item \textbf{Learned positional embeddings:} Transformers, being permutation-invariant, require positional embeddings to assign unique identities to brain regions. These learned embeddings differentiate between regions even with similar activity profiles and encode region-specific biases over training, such as baseline activity or connectivity differences. 
\end{itemize}  

By concatenating these three components, each node representation integrates functional signals (via fMRI), structural topology (via eigenvectors), and regional identity (via positional embeddings), providing the Transformer with rich contextual information to model attention, dynamics, and inter-regional relationships comprehensively. Note that a $[\text{CLS}]$ token is omitted in the Transformer, as this setup focuses on region-wise embeddings without aggregating node information directly from the model.

\subsection{Implementation details}
All deep learning methods were optimized using a combined objective function comprising Mean Absolute Error (MAE), Mean Squared Error (MSE), and negative Pearson correlation. This joint loss ensures accurate behavioral score predictions while maintaining a strong correlation between predicted and true values. The model achieving the best performance on this joint objective in the validation set was selected for evaluation.  

The GNN consists of $[3, 4]$ ChebConv layers, with internal MLP of projection dimension 64, used to refine the node feature representations. 
The MLP regression head includes $[5, 9]$ fully connected layers, each with a hidden dimension of 128, to map graph-level representations to behavioral scores. 
For the Transformer, 2 stacked layers are used to process the input node features. As described in Section \textbf{Transformer-GNN with SC and fMRI Time-series Features}
, the input to the Transformer $\in \mathbb{R}^{N \times K}$ node feature matrix, where $K = X + E + P$. Here, $X$ is the output from an MLP mapping time-series data (Rest: 1194, Language: 310, Working Memory: 399 dimensions) into intermediate dimensions ($256 \rightarrow 64$), $E$ represents the top 16 structural eigenvectors, and $P$ comprises a 16-dimensional learnable positional embedding.

All models were trained for up to 1000 epochs, with early stopping applied if validation performance did not improve for 50 consecutive epochs. Optimization was performed using the AdamW optimizer, with a learning rate sampled from the range $[10^{-4}, \dots, 1]$. A batch size of 32 and dropout values ranging from $[0.05-0.5]$ were used to prevent overfitting during training.

\section{Statistical tests}
Since a 10-fold cross-validation was conducted, we obtained a performance metric for each fold, for each fMRI modality, and for each model. A corrected resampled \textit{t}-test \citep{he2020deep, nadeau1999inference, bouckaert2004evaluating} was used to perform pairwise comparisons: (i) between all model pairs within each fMRI modality, and also across the average predictions from all fMRI modalities; and (ii) between all fMRI modality pairs for each model, as well as across the average predictions from all models. The corrected resampled \textit{t}-test accounts for the dependency between test fold accuracies.

\begin{table}[!t]
    \centering
    \setlength{\tabcolsep}{6pt}
    \small
    \renewcommand{\arraystretch}{0.7}
    \begin{tabular}{l|c|c|c|c|c}
        \toprule
        \multirow{2}{*}{\diagbox[width=11em]{\textbf{fMRI Datasets}}{\textbf{Methods}}} 
        & Kernel Ridge & GNN + & GNN + & GNN + & \multirow{2}{*}{Mean (Methods)} \\
        & Regression & SC + FC & SC + fMRI & SC + TrfMRI & \\
        \midrule
        Rest1 &  0.131 ± 0.087 &  0.160 ± 0.079 &  0.046 ± 0.014 &  0.053 ± 0.042 & 0.098 ± 0.078 \\
        Working Memory &  0.215 ± 0.043 &  0.236 ± 0.061 &  0.142 ± 0.068 &  0.225 ± 0.092 & 0.204 ± 0.076 \\
        Language &  0.243 ± 0.102 &  0.250 ± 0.092 &  0.156 ± 0.054 &  0.225 ± 0.052 & 0.219 ± 0.084 \\
        \midrule
        Mean (fMRI data) & 0.196 ± 0.092 & 0.216 ± 0.086 & 0.114 ± 0.070 & 0.168 ± 0.104 & -- \\
        \bottomrule
    \end{tabular}
    \caption{R$^2$-score (Mean $\pm$ Std) across different fMRI data and competing methods.}
    \label{tab:r2_score}
    \vspace{-4em}
\end{table}

\section{Results}

\subsection{Comparison of fMRI Datasets (Rest, Language, and Working Memory) across Different Methods}

\begin{figure}[!t]
\centerline{\includegraphics[width=0.95\textwidth]{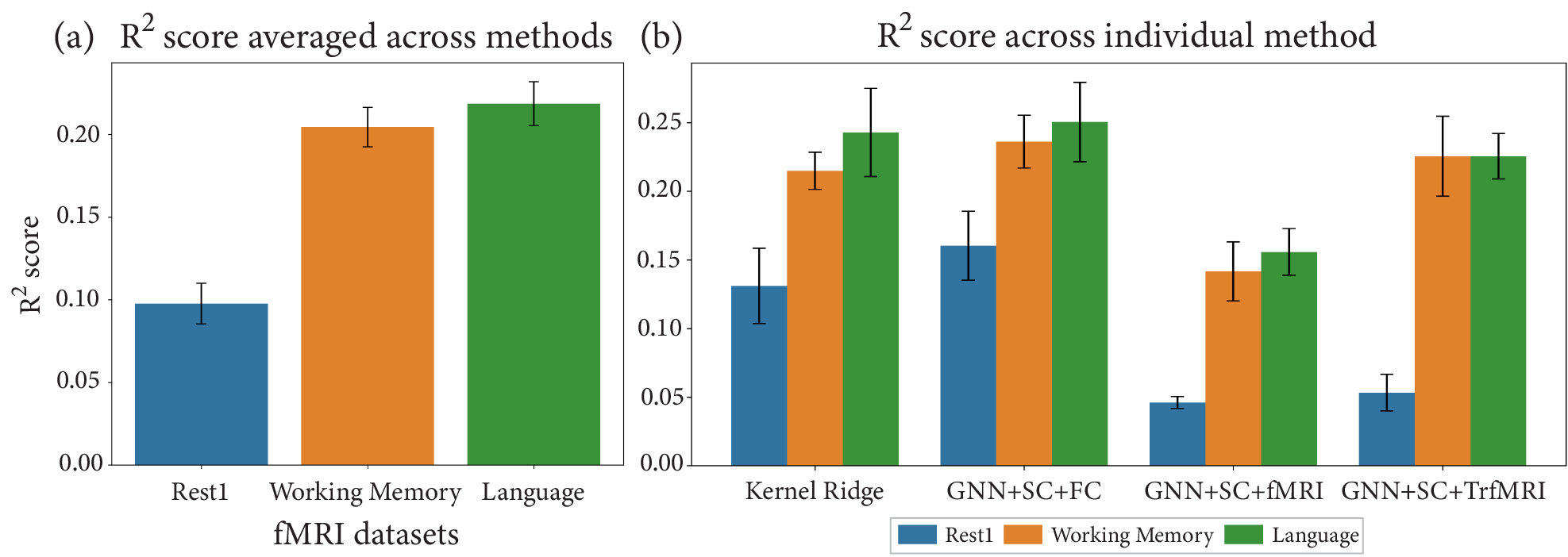}}
\caption{Comparing R$^2$-scores for fMRI datasets (a) averaged across different methods, and (b) across individual method.}
\label{fig:R2method}
\vspace{-2em}
\end{figure}

As shown in \textbf{Table \ref{tab:r2_score}} and \textbf{Figure \ref{fig:R2method} (a)}, the average predictive performance, measured by the R$^2$-score across all methods, is lowest for the resting-state fMRI data (Rest1: $0.098 \pm 0.078$). In contrast, task-based fMRI data (Language and Working Memory) consistently exhibit higher predictive performance. Specifically, the average R$^2$-score for the Language task is slightly higher than for the Working Memory task; however, statistical analysis reveals no significant difference between the two (p-value = 0.487, \textbf{Figure \ref{fig:R2pval} (a)}). Importantly, the average performance of Rest1 is significantly lower than both task-based datasets (p-value $<$ 0.05 for both comparisons).

\textbf{Table \ref{tab:r2_score}} shows that this pattern also holds true across individual methods. For Kernel Ridge Regression (KRR), Rest1 achieves the lowest R$^2$-score ($0.131 \pm 0.087$) compared to Working Memory ($0.215 \pm 0.043$) and Language ($0.243 \pm 0.102$). Statistical analysis confirms a significant difference between Rest1 and Language (p-value = 0.01), while the difference between Rest1 and Working Memory is marginally insignificant (p-value = 0.06). Although Language performs slightly better on average than Working Memory, their difference is not statistically significant.  

The GNN+SC+FC model exhibits similar trends, with Rest1 yielding the poorest performance ($0.160 \pm 0.079$) and Working Memory and Language performing comparably. For this method, no significant difference is observed between the R$^2$-scores of the three types of fMRI datasets.  

The GNN+SC+fMRI model aligns with the overall pattern, showing Rest1 as the lowest-performing fMRI data ($0.046 \pm 0.014$), while Working Memory ($0.142 \pm 0.068$) and Language ($0.156 \pm 0.054$) demonstrate significantly better performance than Rest1 (Rest1 vs. Working Memory: p-value = 0.02; Rest1 vs. Language: p-value = 0.003).  

Similarly, the GNN+SC+TrfMRI model maintains Rest1 as the worst-performing dataset ($0.053 \pm 0.042$), significantly lower than both task-based fMRI data (Rest1 vs. Working Memory: p-value = 0.002; Rest1 vs. Language: p-value = $5 \times 10^{-5}$). Notably, Working Memory and Language tasks exhibit identical average R$^2$-scores ($0.225$), with no significant difference between their performances (p-value = 0.99).  

\begin{figure}[!t]
\centerline{\includegraphics[width=0.95\textwidth]{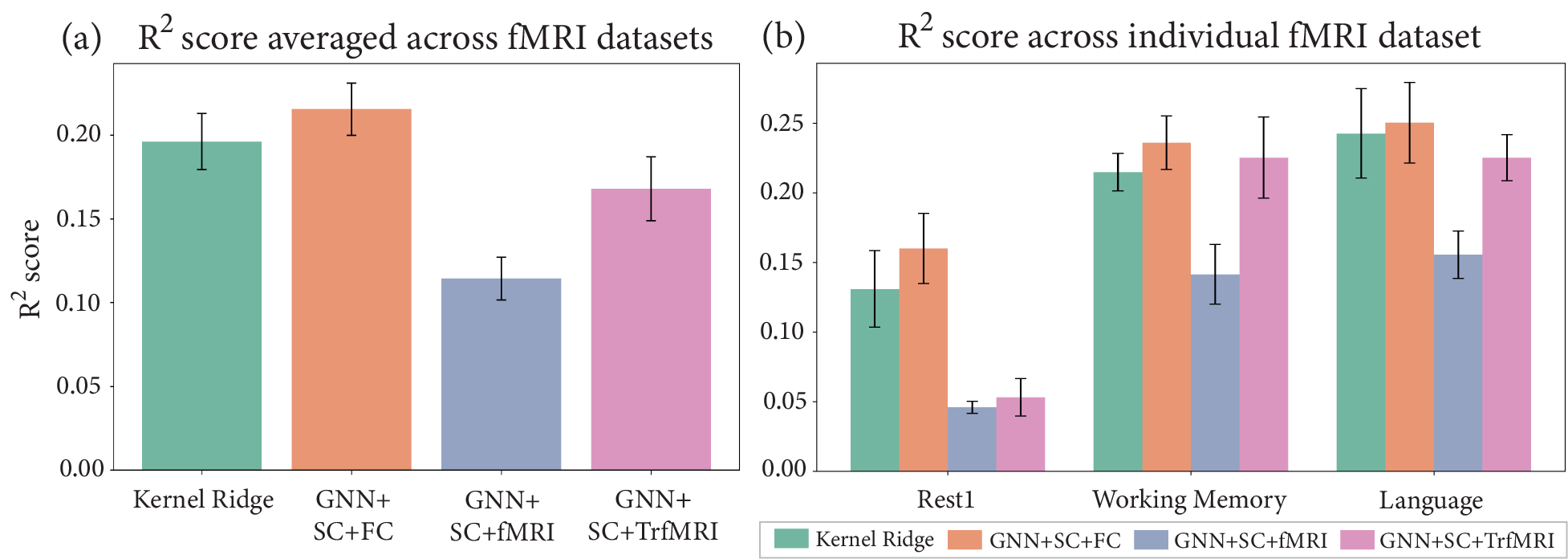}}
\caption{Comparing R$^2$-scores of competing methods (a) averaged across fMRI datasets, and (b) across individual fMRI dataset.}
\label{fig:R2fMRI}
\vspace{-3em}
\end{figure}

\subsection{Comparison of Different Methods across fMRI Datasets (Rest, Language, and Working Memory)}

\begin{figure}[!t]
\centerline{\includegraphics[width=0.95\textwidth]{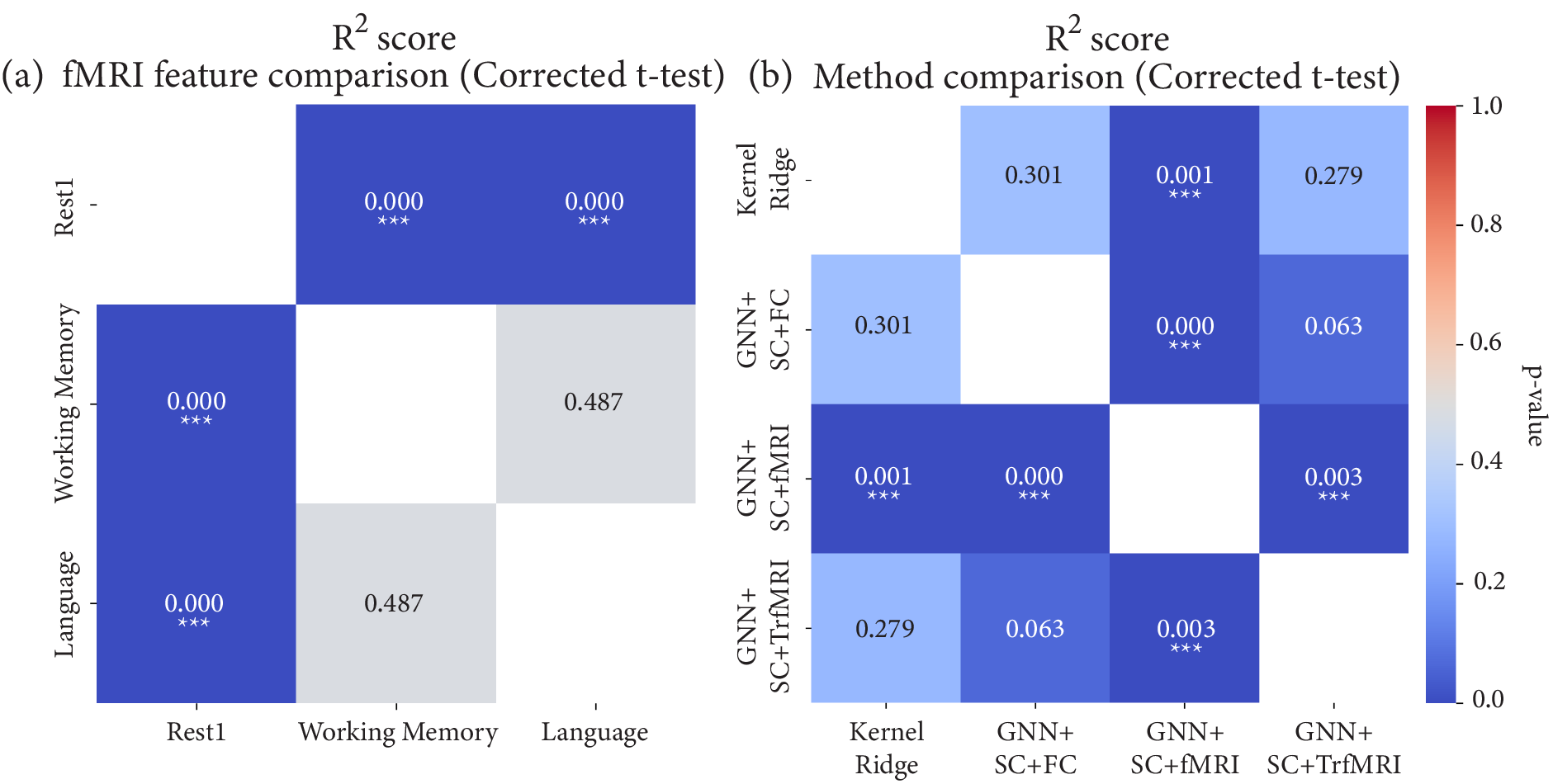}}
\caption{Corrected p-values for pairwise comparisons of (a) fMRI datasets based on R$^2$-scores averaged across methods, and (b) methods based on R$^2$-scores averaged across fMRI datasets.}
\label{fig:R2pval}
\vspace{-2em}
\end{figure}

As illustrated in \textbf{Table \ref{tab:r2_score}} and \textbf{Figure \ref{fig:R2fMRI} (a)}, the average predictive performance across all fMRI datasets, measured by the R$^2$-score, is lowest for GNN+SC+fMRI ($0.114 \pm 0.070$). This performance is significantly lower than that of the other three methods. Among the remaining methods, GNN+SC+FC achieves the highest average R$^2$-score ($0.216 \pm 0.086$), slightly outperforming KRR ($0.196 \pm 0.092$), though this difference is not statistically significant (p-value = 0.301). The average performance of GNN+SC+TrfMRI ($0.168 \pm 0.104$) is lower than both GNN+SC+FC and KRR, but these differences are not statistically significant—for KRR (p-value = 0.279) and approaching significance for GNN+SC+FC (p-value = 0.063) \textbf{Figure \ref{fig:R2pval} (b)}.

\textbf{Figure \ref{fig:R2fMRI} (b)} and \textbf{Table \ref{tab:r2_score}} further highlight method-specific trends across individual fMRI datasets. GNN+SC+FC consistently achieves the highest R$^2$-scores, while GNN+SC+fMRI frequently exhibits the lowest performance. The relative performance of KRR and GNN+SC+TrfMRI depends on the specific fMRI dataset.  

For the Rest1 fMRI data, GNN+SC+TrfMRI performs slightly better than GNN+SC+fMRI, achieving a mean R$^2$-score of $0.053$ compared to $0.046$, though this difference is not statistically significant (p-value = 0.73). Notably, GNN+SC+FC achieves numerically better performance than KRR ($0.16$ vs. $0.13$), but again, the difference is not significant (p-value = 0.49). Methods incorporating FC derived from fMRI data generally outperform those using fMRI data either given directly or processed by a transformer. For Rest1, GNN+SC+FC significantly outperforms both GNN+SC+fMRI (p-value = 0.02) and GNN+SC+TrfMRI (p-value = 0.01).  

For the Working Memory fMRI data, GNN+SC+FC achieves the best average R$^2$-score ($0.236$), followed by GNN+SC+TrfMRI ($0.225$) and KRR ($0.215$), while GNN+SC+fMRI performs the worst ($0.140$). Here, the top three methods perform comparably, with no significant differences among them. However, both GNN+SC+FC (p-value = 0.03) and GNN+SC+TrfMRI (p-value = 0.01) show statistically significant improvements over GNN+SC+fMRI.  

For the Language fMRI data, the highest average R$^2$-score is achieved by GNN+SC+FC ($0.25$), followed by KRR ($0.243$) and GNN+SC+TrfMRI ($0.225$), with GNN+SC+fMRI again performing the worst ($0.156$). Similar to Working Memory, the top three methods show no statistically significant differences among themselves. However, GNN+SC+TrfMRI significantly outperforms GNN+SC+fMRI (p-value = 0.04).

\begin{figure}[!t]
\centerline{\includegraphics[width=0.9\textwidth]{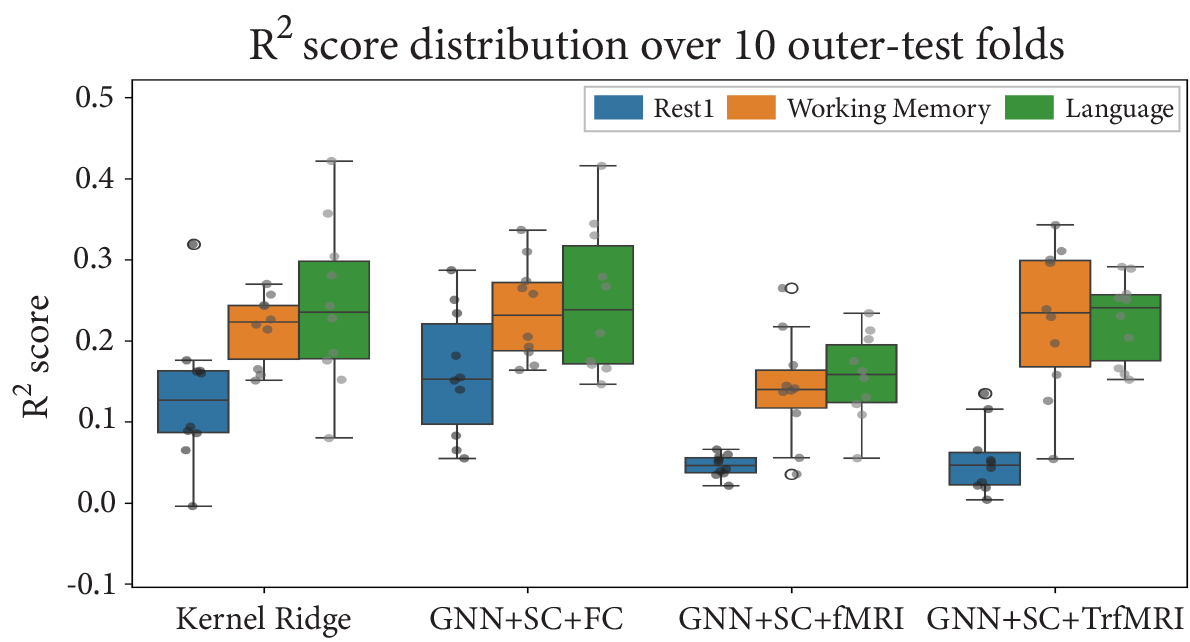}}
\caption{Distribution of R$^2$-scores across 10 outer-test folds during nested cross-validation. The box-plots are presented across three fMRI datasets and four evaluated methods.}
\label{fig:R2box}
\vspace{-2em}
\end{figure}

\begin{figure}[!t]
\centerline{\includegraphics[width=0.95\textwidth]{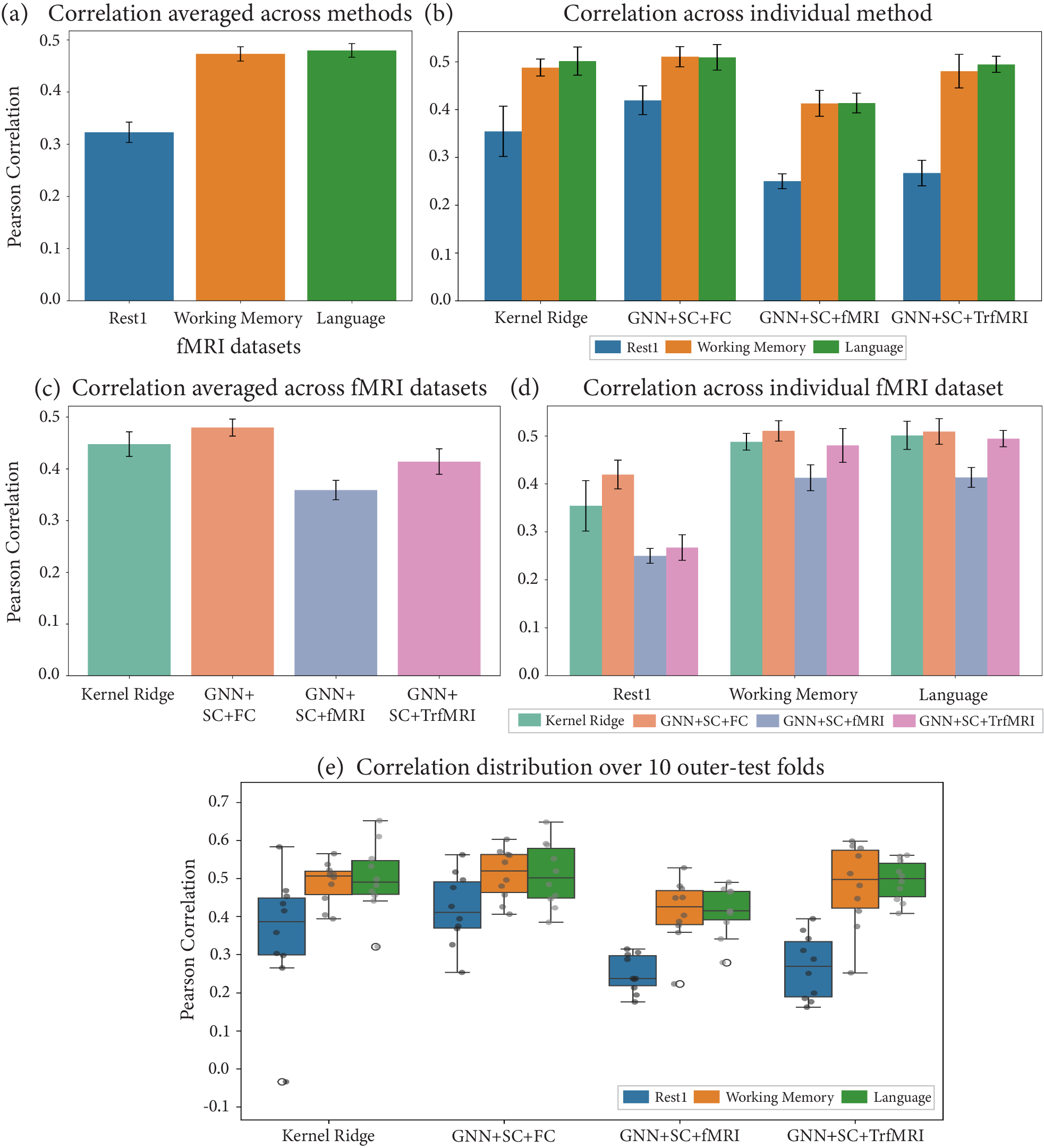}}
\caption{Results for Pearson correlation coefficient (PCC) metric. Comparing PCC for fMRI datasets (a) averaged across different methods, and (b) across individual method. Comparing PCC of competing methods (c) averaged across fMRI datasets, and (d) across individual fMRI dataset. (e) Distribution of PCC across 10 outer-test folds during nested cross-validation. The box-plots are presented across three fMRI datasets and four evaluated methods.}
\label{fig:correlation_results}
\vspace{-2em}
\end{figure}

\begin{figure}[!t]
\centerline{\includegraphics[width=0.95\textwidth]{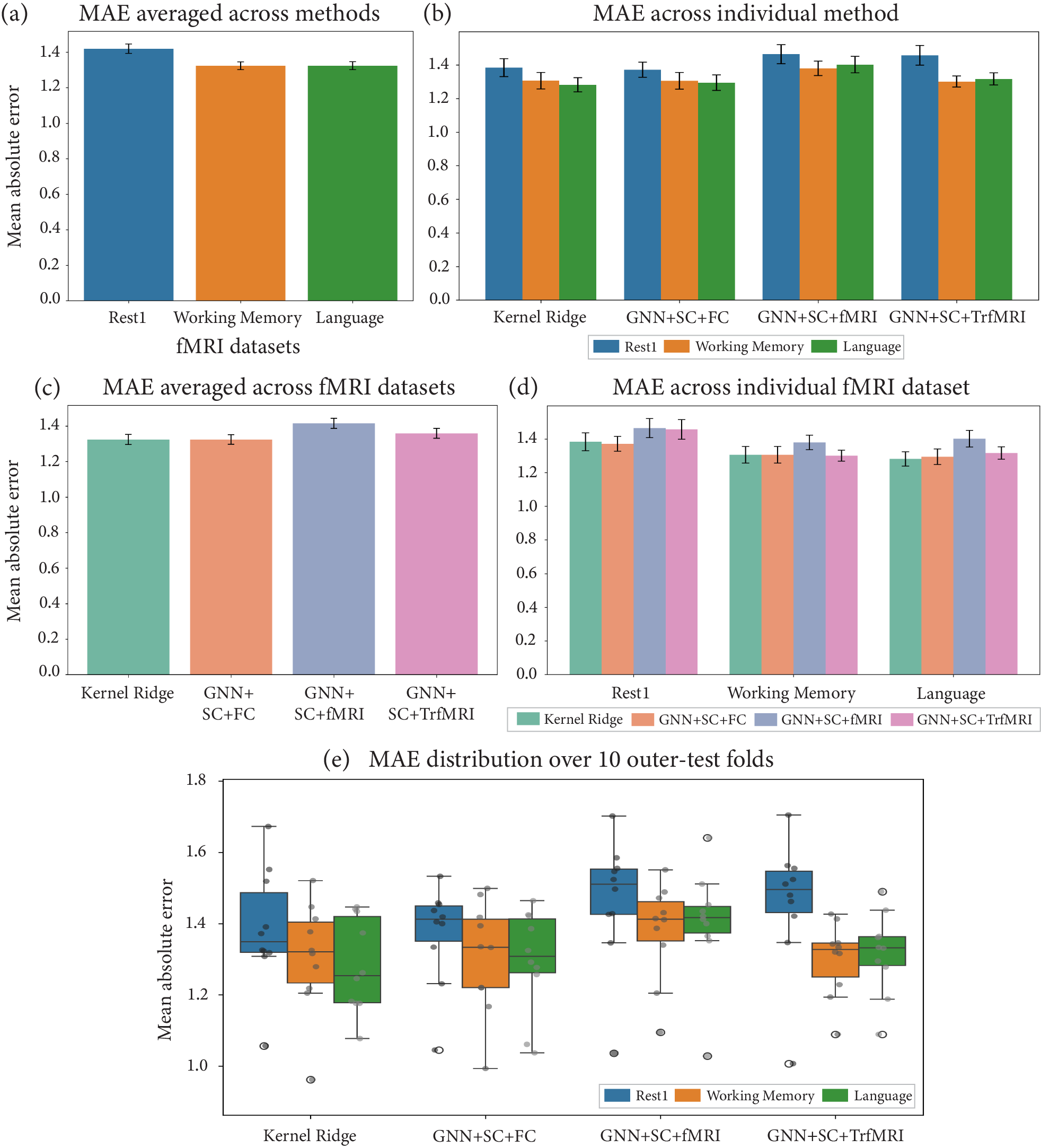}}
\caption{Results for mean absolute error (MAE) metric. Comparing MAE for fMRI datasets (a) averaged across different methods, and (b) across individual method. Comparing MAE of competing methods (c) averaged across fMRI datasets, and (d) across individual fMRI dataset. (e) Distribution of MAE across 10 outer-test folds during nested cross-validation. The box-plots are presented across three fMRI datasets and four evaluated methods.}
\label{fig:mae_results}
\vspace{-2em}
\end{figure}

A detailed visualization of the previously summarized R$^2$-scores across outer-test folds during nested cross-validation is presented in \textbf{Figure \ref{fig:R2box}}, with box-plots depicting results across all evaluated fMRI datasets and methodologies. Additional results for Pearson correlation coefficient (PCC) and mean absolute error (MAE) metrics are shown in \textbf{Figure \ref{fig:correlation_results}} and \textbf{Figure \ref{fig:mae_results}}, respectively. Consistent with R$^2$-scores evaluations, we report  performance averaged across methods, individual methods, averaged across fMRI datasets, and individual fMRI datasets for PCC and MAE. Observations for PCC and MAE metrics align with the trends seen R$^2$-scores.

\section{Discussion}

This study benchmarks machine learning approaches, from Kernel Ridge Regression (KRR) to advanced Graph Neural Networks (GNN)-based approaches, integrating structural connectivity (SC), functional connectivity (FC), and temporal dynamics via a Transformer architecture, for cognitive prediction using Resting-state, Working Memory, and Language task fMRI data from the Human Connectome Project Young Adult (HCP-YA) dataset. Our results highlight key insights into the relative effectiveness of these methods and the importance of leveraging various neuroimaging datasets.

Our findings indicate that Resting-state fMRI is the least effective modality (out of the chosen fMRI modalities) for predicting cognition, as measured by $R^2$-scores across different methods, consistent with prior studies \citep{ooi2022comparison}. This could be attributed to resting-state FC primarily reflecting baseline functional organization, which may lack specificity to cognitive processes compared to task-evoked activity. In contrast, both Working Memory and Language task-based fMRI demonstrated superior predictive performance. Although Language fMRI showed numerically higher predictive scores than Working Memory, this difference was not statistically significant. 
Based on the observations, it can be hypothesized that there exists a potential stronger correlation between the neural activity elicited during the language task in the HCP dataset and the behavioral measures contributing to the primary cognition factor, compared to the working memory task. However, this observation requires validation across larger datasets and a broader range of cognitive assessments.
Nevertheless, the comparable predictive power observed in this study underscores that both Working Memory and Language paradigms elicit informative neural activity patterns for cognitive prediction.

Noticeably, our KRR results were lower than those reported by \citep{ooi2022comparison}, despite using the same cognitive scores. This discrepancy could stem from several methodological differences. We excluded global signal regression during fMRI preprocessing, a step shown to improve behavior prediction in Resting-state fMRI \citep{li2019global}. We also employed a parcellation of 274 brain regions, whereas \citep{ooi2022comparison} used a finer-grained parcellation of 400 regions. Moreover, we utilized only one run of Rest1 and one run for each task, while \citep{ooi2022comparison} incorporated two runs for both Resting-state (Rest1 and Rest2) and task-based fMRI. Prior research also suggests that increased scan duration enhances the predictive power of cognition \citep{ooi2024mri}, which could further contribute to the observed differences.

Among all methods, GNNs integrating structural and functional connectivity (GNN+SC+FC) consistently achieved the best predictive performance, both when averaged across fMRI datasets and for individual fMRI data. This aligns with prior work \citep{xia2025interpretable,kramer2024prediction} and underscores the complementary strengths of SC and FC for cognition prediction: SC defines the brain’s physical network, while FC captures regional functional correlations. Notably, the comparable performance of Rest1 and task-based data for this method (statistically no significant difference) suggests that functional activations during rest might be more directly guided by the underlying structural connections compared to the more task-modulated activity. This stronger influence of structure on resting-state function could be a reason why the model performs relatively well even with resting-state data, potentially capturing fundamental structural-functional relationships relevant to cognition that are less pronounced or more variable during task engagement.

In contrast, GNNs using SC as structural priors with direct fMRI time-series data (GNN+SC+fMRI) consistently performed the worst across all fMRI datasets. While prior work \citep{vieira2024beyond} demonstrated comparable performance between KRR on FC and temporal CNNs on fMRI time-series, our results suggest that GNNs are less effective at capturing temporal patterns directly from time-series data. Preprocessing time-series with temporal CNNs at the regional level before applying graph-based models, as done by \citep{vieira2024beyond}, may better extract spatial-temporal features and improve performance.  

The spatio-temporal Transformer-GNN model (GNN+SC+TrfMRI) exhibited mixed results. Although it performed comparably to GNN+SC+FC on task-based fMRI, its performance on resting-state data was poor, comparable to GNN+SC+fMRI. This may reflect the Transformer’s sensitivity to input dimensionality ($\sim1200$ time points) and relevance, as its region-wise embedding of Resting-state scans may not adequately capture the low-frequency dynamics, unlike task-based scans with fewer but more behaviorally specific time points. This observation indicates that more sophisticated temporal feature extraction, beyond MLPs, are required to effectively capture the low-frequency dynamics. Additionally, while the FC derived from the Transformer-based embeddings was competitive with static FC for task data, this approach may require further optimization to better combine temporal encoding and SC for cognition prediction. Nevertheless, GNN+SC+TrfMRI demonstrated performance comparable to GNN+SC+FC, highlighting the potential of Transformer-based architectures to capture data-driven temporal interactions beyond the static correlations modeled in FC. With larger datasets and further optimization, this approach holds promise as a powerful future direction for cognition prediction.  

Comparing the deep learning methods with classical machine learning, KRR provided a strong baseline for predicting cognition, particularly for task-based fMRI, and its performance was competitive with more advanced models. Although deep learning approaches like GNN+SC+FC generally outperformed KRR, this difference was not statistically significant. Also, it remains unclear whether the improvement stems from the incorporation of SC or the potential of deep neural networks like GNN, requiring further investigation. This aligns with prior findings \citep{he2020deep} that simple machine learning models can perform comparably to deep learning approaches for behavior prediction from FC. Limited sample size, model complexity, and extensive hyperparameter tuning requirements for deep learning models may explain the lack of significant improvement \citep{he2020deep,yeung2023predicting}. As showcased by \citep{yeung2023predicting}, even 8000 samples might be insufficient to fully leverage the abilities of deep networks for behavioral or demographic prediction. Our results indicate that while the non-linear modeling capabilities of deep learning, particularly when effectively integrating multimodal information like SC and FC, can offer an advantage over classical methods on the HCP-YA dataset, the choice of architecture and feature representation emerges as a critical factor in achieving optimal performance.  

Examining model performance across fMRI datasets revealed distinct patterns, highlighting that different methods excel at capturing unique brain-behavior relationships depending on the fMRI modality. This variability underscores the importance of modality-specific optimization to tailor models for effectively capturing the neural correlates of particular cognitive domains.

\section{Limitations and Future Work}

This study provides valuable insights but comes with certain limitations. First, the use of the high-quality HCP-YA dataset, restricted to a narrow age range (22–37 years), limits the generalizability of findings to datasets with broader demographic diversity, wider age spectrums, and varying imaging protocols. 
Second, the comparison between Kernel Ridge Regression with FC and GNNs incorporating both SC and FC raises the unresolved question of whether the observed improvement is due to the inclusion of structural knowledge or the learning ability of the GNN architecture. Evaluating GNN performance on FC alone versus the multimodal GNN could disentangle the contributions of GNN modeling and multimodality.  
Third, the finding that FC derived from Transformer-learned embeddings was as informative as static FC for task-fMRI prompts further investigation into whether this result is limited by sample size or the integration strategy for temporal information with SC in GNNs. Exploring alternative Transformer architectures and improved strategies for fusing temporal features with structural connectivity could better exploit the rich temporal dynamics of fMRI data. Additionally, given the complexity of deep learning architectures, extensive hyperparameter optimization is necessary to ensure models are fully trained and perform at their best.  

Future work could also benefit from incorporating finer-grained brain parcellations, modeling all task-based fMRI data jointly, and including other SC edge weights, such as fractional anisotropy or mean diffusivity. Extending the analysis to additional behavioral domains, including dissatisfaction or emotional processing \citep{ooi2022comparison}, could broaden the scope of these findings by uncovering neural correlates for diverse behavioral dimensions. Finally, enhancing the interpretability of these complex models is critical for uncovering the underlying neural mechanisms driving predictions. Leveraging techniques like attention maps or feature importance analysis could help identify the specific brain regions and connectivity patterns most predictive of cognitive behavior.

\section{Conclusion}

This study benchmarked classical machine learning (Kernel Ridge Regression) against advanced deep learning models, including GNNs and spatio-temporal Transformer-GNNs, for cognition prediction using fMRI data. While Kernel Ridge Regression established a strong baseline for prediction with FC, the superior performance of the SC+FC+GNN model highlights the effectiveness of multimodal approaches that combine structural and functional connectivity using graph-aware networks. The mixed results from the Transformer-GNN model and the lower performance of GNNs utilizing direct time-series data emphasize the challenges of fully capturing the spatio-temporal complexity of fMRI data. These findings underline the need for further methodological innovations in model architectures and feature integration strategies to harness the rich temporal and spatial dynamics of fMRI data for accurate and robust behavioral prediction. Future research should prioritize advancing deep learning models and multimodal approaches to improve behavior prediction and benchmark on larger sample sets, ultimately deepening our understanding of brain-behavior relationships across both healthy and clinical populations.  

\newpage

\newpage

\newpage

\newpage


\bibliography{bibliography}

\end{document}